\documentclass[letterpaper]{article} 
\usepackage{aaai2026}  
\usepackage{times}  
\usepackage{helvet}  
\usepackage{courier}  
\usepackage[hyphens]{url}  
\usepackage{graphicx} 
\usepackage{natbib}  
\usepackage{caption} 
\usepackage{algorithm}
\usepackage{algorithmic}

\usepackage{amsfonts} 
\usepackage{enumitem}
\usepackage{amsmath}
\usepackage{amsthm}
\usepackage{multirow}
\usepackage{graphicx}
\usepackage{pifont}
\usepackage{booktabs}
\usepackage{subcaption} 
\newtheorem{theorem}{Theorem}

\usepackage{newfloat}
\usepackage{listings}
\DeclareCaptionStyle{ruled}{labelfont=normalfont,labelsep=colon,strut=off} 
\floatstyle{ruled}
\newfloat{listing}{tb}{lst}{}
\floatname{listing}{Listing}
\title{DcMatch: Unsupervised Multi-Shape Matching with Dual-Level Consistency}
\author{
    Tianwei Ye, Yong Ma, Xiaoguang Mei\thanks{Corresponding author.}
}
\affiliations{
    Electronic Information School, Wuhan University, Wuhan 430072, China \\

    twye2001@gmail.com, mayong@whu.edu.cn, meixiaoguang@whu.edu.cn
}

\usepackage{bibentry}

\begin{document}

\maketitle

\begin{abstract}
Establishing point-to-point correspondences across multiple 3D shapes is a fundamental problem in computer vision and graphics. In this paper, we introduce DcMatch, a novel unsupervised learning framework for non-rigid multi-shape matching. Unlike existing methods that learn a canonical embedding from a single shape, our approach leverages a shape graph attention network to capture the underlying manifold structure of the entire shape collection. This enables the construction of a more expressive and robust shared latent space, leading to more consistent shape-to-universe correspondences via a universe predictor. Simultaneously, we represent these correspondences in both the spatial and spectral domains and enforce their alignment in the shared universe space through a novel cycle consistency loss. This dual-level consistency fosters more accurate and coherent mappings. Extensive experiments on several challenging benchmarks demonstrate that our method consistently outperforms previous state-of-the-art approaches across diverse multi-shape matching scenarios.
\end{abstract}

\begin{links}
    \link{Code}{https://github.com/YeTianwei/DcMatch}
\end{links}

\section{Introduction}
Shape matching is a fundamental problem in computer vision and graphics \cite{van2011survey, sahilliouglu2020recent}, which aims to establish accurate point-to-point correspondences between 3D shapes. It supports a wide range of applications, including texture transfer in graphics, statistical shape analysis in medical imaging, and 3D reconstruction in computer vision. While most existing work focuses on matching shape pairs, recent advances in 3D scanning technology have made it increasingly common to capture multiple shapes simultaneously—for instance, different scans of the same object under varying conditions. In this context, establishing correspondences across a collection of shapes, known as multi-shape matching, becomes essential.

Although recent pairwise shape matching methods have achieved promising results \cite{ bastian2024hybrid, zhuravlev2025denoisfm}, extending them to the multi-shape setting poses significant challenges. First, multi-shape matching requires cycle consistency across the shape collection—namely, the composition of maps along any closed cycle should yield the identity map. This global constraint, which is absent in pairwise settings, considerably increases the problem's complexity. Secondly, the number of shape pairs grows combinatorially with the size of the shape set, leading to substantial computational overhead.

\begin{figure}[t]
\centering
\includegraphics[width=1\linewidth]{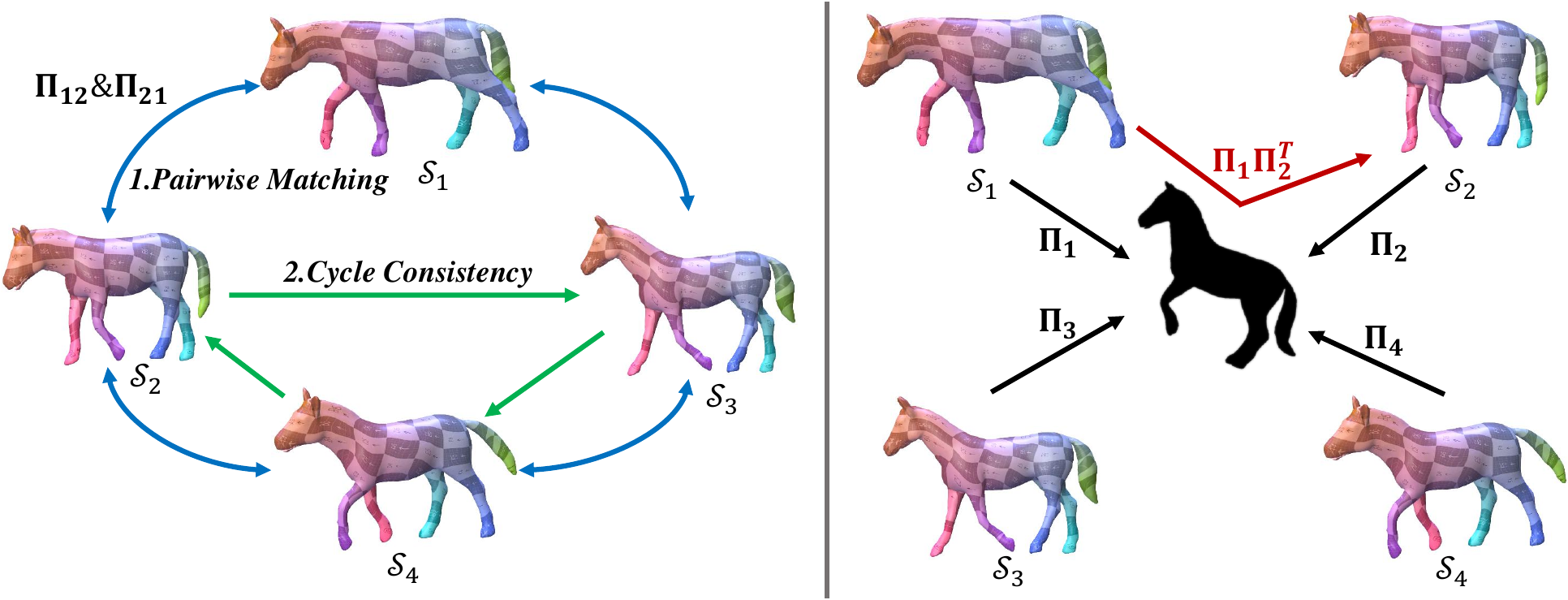}
\caption{
    \textbf{Overview of mainstream multi-shape matching paradigms.} (Left) The permutation synchronization paradigm, which consists of two stages: computing pairwise correspondences and enforcing cycle consistency via post-processing. (Right) The universe-based paradigm, which introduces a virtual universe shape and reduces the multi-shape matching problem to a set of pairwise mappings.}
\label{fig:formulation}
\end{figure}

To address the challenges of multi-shape matching, existing methods generally fall into two main paradigms, as illustrated in Fig.~\ref{fig:formulation}. The first paradigm imposes cycle consistency constraints directly on pairwise mappings to enforce global consistency. Known as permutation synchronization \cite{pachauri2013solving, bernard2019synchronisation}, this approach  offers flexibility but typically involves a two-stage optimization process and often leads to spatially non-smooth and noisy results. The second paradigm introduces a shared latent domain, commonly referred to as the \emph{universe}, which converts pairwise correspondences into mappings between each shape and the universe \cite{cosmo2017consistent, bernard2019hippi}. This formulation enforces global consistency by construction. However, most methods within this paradigm typically learn the universe embedding from a single shape, in either the spatial \cite{cao2022unsupervised} or spectral \cite{huang2013consistent} domain, thereby reducing multi-shape matching to a collection of isolated pairwise problems. This often neglects the structural relationships within the shape collection and leads to suboptimal performance \cite{eisenberger2023g}.

To tackle the above limitations, we propose DcMatch, a novel unsupervised multi-shape matching framework that enforces dual-level cycle consistency. Our pipeline begins by extracting per-vertex features for each shape using a feature extractor. Inspired by the accuracy of pairwise matching in permutation synchronization methods, we directly compute functional maps and point-wise correspondences from these features. In parallel, instead of learning a canonical embedding from a single shape as in prior universe-based approaches, we model the entire shape collection as an undirected graph and employ a graph attention network to extract manifold-aware features that capture the underlying geometric structure. These features are then fed into a universe predictor to estimate shape-to-universe correspondences. By doing so, our model effectively learns a universe embedding enriched with manifold information. Beyond this, by leveraging the inherent cycle consistency of functional maps combined with the global consistency of shape-to-universe matching, we enforce consistency between these two alignment paths, thereby improving robustness and consistency in the shared universe space. Through this dual-level consistency, our method achieves more accurate and coherent multi-shape matching in a fully unsupervised setting. Extensive experiments on diverse datasets demonstrate the competitive performance of our approach.

We summarize our main contributions as follows:
\begin{enumerate}
\item We propose a novel unsupervised framework for multi-shape matching that enforces both spectral and spatial cycle consistency, resulting in more accurate and coherent correspondence predictions.
\item We introduce a shape graph attention module that captures the underlying manifold structure of the entire shape collection. This facilitates the construction of a robust and generalizable universe space, in contrast to prior methods that learn from a single reference shape.
\item We conduct extensive experiments under various settings, demonstrating that our method achieves state-of-the-art performance across challenging benchmarks.
\end{enumerate}

\section{Related Work}
We refer interested readers to \cite{van2011survey, sahilliouglu2020recent} for a comprehensive review. Here, we highlight only the approaches most relevant to our method.

\subsection{Functional Maps}
Functional maps, first introduced by \cite{ovsjanikov2012functional}, have become one of the most widely used frameworks in shape matching. By representing point-wise correspondences as compact matrices in the spectral domain, they offer substantial computational efficiency. Thanks to their concise and flexible formulation, functional maps have been extended in various directions to improve accuracy and robustness \cite{eynard2016coupled, melzi2019zoomout, ren2019structured}, and to handle more challenging scenarios, including partial \cite{litany2017fully, rodola2017partial}, non-isometric \cite{nogneng2017informative, ren2021discrete, ren2018continuous}, and non-unique shape matching \cite{ren2020maptree}.

\subsection{Learning Method Based on Functional Maps}
Unlike axiomatic functional map methods that depend on hand-crafted descriptors \cite{sun2009concise, aubry2011wave, salti2014shot}, recent approaches learn feature descriptors directly from data. FMNet \cite{litany2017deep} introduced a supervised framework that transforms SHOT descriptors into more effective embeddings for functional map estimation. This was extended to the unsupervised setting in \cite{halimi2019unsupervised, roufosse2019unsupervised} by introducing regularization-based losses. More recently, DiffusionNet \cite{sharp2022diffusionnet} has enabled robust, resolution-aware feature extraction, inspiring several state-of-the-art methods \cite{cao2022unsupervised, cao2023unsupervised, li2022learning, donati2022deep, bastian2024hybrid} that perform well across diverse shape datasets.

\subsection{Multi-shape Matching}
Early multi-shape matching methods enforce cycle consistency via semidefinite programming or convex relaxation, but often yield sparse correspondences and poor scalability. Within the functional map framework, Consistent ZoomOut \cite{huang2020consistent} synchronizes maps via a shared basis but depends on good initialization. IsoMuSh \cite{gao2021isometric} jointly enforces consistency on pointwise and functional maps but is restricted to near-isometric shapes. CycoMatch \cite{xia2025multi} builds cycle-consistent bases via a graph in a two-stage pipeline. Recently, learning-based approaches have improved scalability. UDMSM \cite{cao2022unsupervised} predicts a canonical embedding to promote cycle consistency, while G-MSM \cite{eisenberger2023g} leverages a heuristic to model the shape collection’s underlying manifold.

\section{Background}
In this section, we first review the deep hybrid functional maps pipeline \cite{bastian2024hybrid}, then revisit cycle consistency with corresponding theoretical insights.

\subsection{Deep Hybrid Functional Maps}
Given a pair of 3D shapes $\mathcal{S}_i$ and $\mathcal{S}_j$, represented as triangle meshes with $n_i$ and $n_j$ vertices respectively, the hybrid functional map framework aims to represent dense correspondences in a compact, linear form. The main pipeline consists of the following steps:

\begin{enumerate}
\item Compute two sets of basis functions for each shape: (i) the first $k_{\mathrm{LB}}$ eigen-functions of the Laplace Beltrami operator (LBO) \cite{pinkall1993computing}, denoted as $\Phi_i \in \mathbb{R}^{n_i \times k_{\mathrm{LB}}}$, and (ii) the first $k_{\mathrm{Elas}}$ eigen-functions of the elastic thin-shell energy \cite{hartwig2023elastic}, denoted as $\Psi_i \in \mathbb{R}^{n_i \times k_{\mathrm{Elas}}}$.

\item Extract vertex-wise feature descriptors $\mathcal{F}_i \in \mathbb{R}^{n_i \times d}$ using a scalable network \cite{sharp2022diffusionnet}, where $d$ is the feature dimension. These features are projected onto both sets of basis functions to obtain spectral coefficients $\mathcal{A}^{\mathrm{LB}}_i = \Phi_i^\dagger \mathcal{F}_i \in \mathbb{R}^{k_{\mathrm{LB}} \times d}$, $\mathcal{A}^{\mathrm{Elas}}_i = \Psi_i^\dagger \mathcal{F}_i \in \mathbb{R}^{k_{\mathrm{Elas}} \times d}$. The combined hybrid basis is denoted as $\widetilde{\Phi}_i = [\Phi_i ; \Psi_i]$, with corresponding projected features $\mathcal{A}_i = \widetilde{\Phi}_i^\dagger \mathcal{F}_i \in \mathbb{R}^{(k_{\mathrm{LB}} + k_{\mathrm{Elas}}) \times d}$.

\item Solve the block-diagonal map $C_{ij} = \mathrm{diag}(C^{11}_{ij}, C^{22}_{ij}) \in \mathbb{R}^{(k_{\mathrm{LB}} + k_{\mathrm{Elas}}) \times (k_{\mathrm{LB}} + k_{\mathrm{Elas}})}$ by minimizing regularized objectives per basis:
\begin{equation}
\begin{gathered}
C^{11}_{ij} = \arg\min_C E^{\mathrm{LB}}_{\text{data}}(C) + \lambda_{\mathrm{LB}} E^{\mathrm{LB}}_{\text{reg}}(C), \\
E^{\mathrm{LB}}_{\text{data}}(C) = \left\| C \mathcal{A}^{\mathrm{LB}}_i - \mathcal{A}^{\mathrm{LB}}_j \right\|_{\mathrm{F}}^2, \\
E^{\mathrm{LB}}_{\text{reg}}(C) = \left\| C \Lambda^{\mathrm{LB}}_i - \Lambda^{\mathrm{LB}}_j C \right\|_{\mathrm{F}}^2,
\end{gathered}
\label{eq:LB functional map}
\end{equation}
\begin{equation}
\begin{gathered}
C^{22}_{ij} = \arg\min_C E^{\mathrm{Elas}}_{\text{data}}(C) + \lambda_{\mathrm{Elas}} E^{\mathrm{Elas}}_{\text{reg}}(C), \\
 E^{\mathrm{Elas}}_{\text{data}}(C) = \left\| C \mathcal{A}^{\mathrm{Elas}}_i - \mathcal{A}^{\mathrm{Elas}}_j \right\|_{M_{k_\mathrm{Elas},j}}^2, \\
 E^{\mathrm{Elas}}_{\text{reg}}(C) =  \left\| C \Lambda^{\mathrm{Elas}}_i - \Lambda^{\mathrm{Elas}}_j C \right\|_{\mathrm{HS}}^2,
\end{gathered}
\label{eq:elas functional map}
\end{equation}
where ${M_{k_\mathrm{Elas},j}} = \Psi_j^\top M_j \Psi_j$ denotes the reduced mass matrix induced by the elastic basis on shape $\mathcal{S}_j$, $\left\| \cdot \right\|_{\mathrm{HS}}$ represents the Hilbert-Schmidt norm and $\Lambda_i, \Lambda_j$ are diagonal eigenvalue matrices for each basis.

\item Convert the hybrid functional map $C_{ij}$ into a point-wise correspondence $\Pi_{ji} \in \{0,1\}^{n_j \times n_i}$ using nearest-neighbor search or other post-processing techniques \cite{melzi2019zoomout, pai2021fast, vestner2017product, xia2024locality}, based on the relation:
\begin{equation}
\widetilde{\Phi}_j C_{ij} \approx \Pi_{ji} \widetilde{\Phi}_i.
\end{equation}
\end{enumerate}

\begin{figure*}[t]
\centering
\includegraphics[width=1\linewidth]{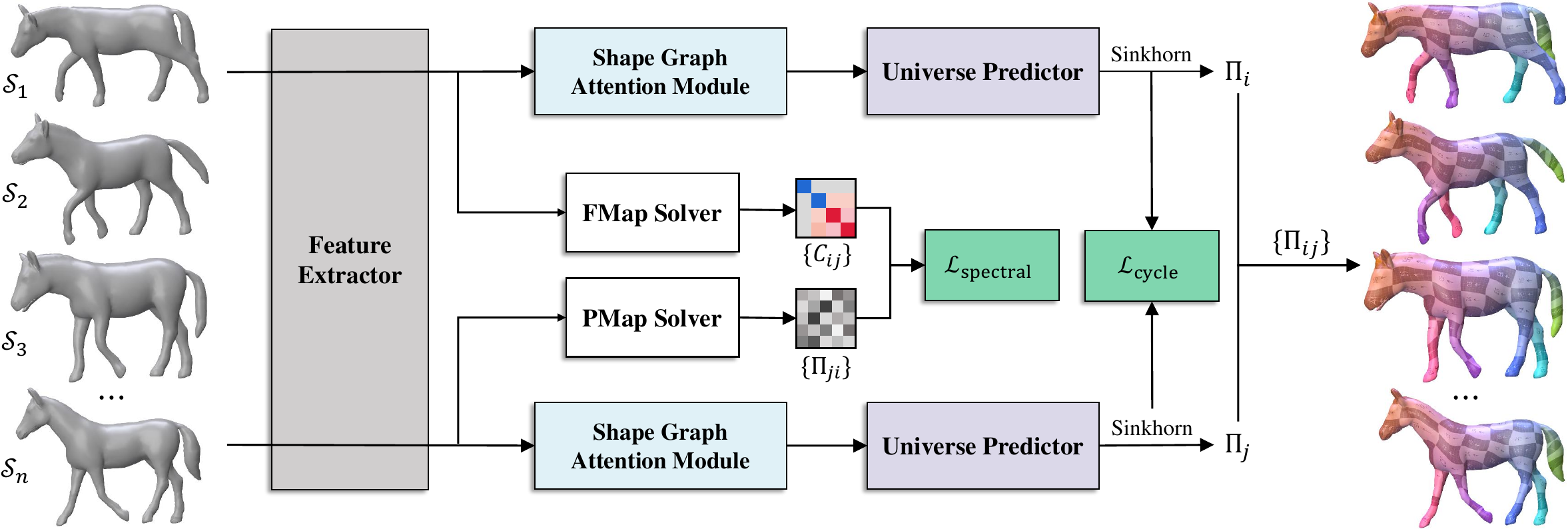}
\caption{
    \textbf{Method Overview.} Given a collection of shapes $\mathcal{S} = \{\mathcal{S}_i\}_{i=1}^n$, we first extract per-vertex features $\mathcal{F} = \{\mathcal{F}_i \} _{i=1}^n$ using DiffusionNet. These features are then used to compute bidirectional functional maps $\{C_{ij}\}$ and point-to-point correspondences $\{\Pi_{ij}\}$. Meanwhile, the shape graph attention module generates manifold-aware features for each shape, which are passed to the universe predictor to estimate the correspondences between each shape and a shared virtual universe. In addition to the spectral loss, we implement a cycle consistency loss to further align the spatial and spectral consistency in the shared universe space.
}
\label{fig:framework}
\end{figure*}

\subsection{Cycle Consistency Formulation}
\label{subsec:cycle_consistency}
We begin by defining the notion of cycle consistency. Let $\mathcal{S} = \{ \mathcal{S}_i\}_{i=1}^{n}$ be a collection of shapes, and let $\mathcal{P} = \{\Pi_{ij}\}_{i,j=1}^{n}$ and $\mathcal{C} = \{C_{ij}\}_{i,j=1}^{n}$ denote the sets of point-wise and functional maps, respectively. The set $\mathcal{P}$ is said to be cycle consistent if, for any closed sequence of shapes $\{i, j, k, \dots, l, i\}$, the composed map satisfies $\Pi_{ij} \Pi_{jk} \dots \Pi_{li} = \mathbf{I}$, where $\mathbf{I}$ denotes the identity map \cite{huang2013consistent}. A similar definition applies to the set of functional maps $\mathcal{C}$.

Rather than explicitly enforcing cycle consistency over all shape pairs, the same principle can be realized implicitly through a \emph{shape-to-universe} formulation \cite{huang2013consistent, tron2017fast, gao2021isometric}. In this setting, each shape is associated with a shared canonical domain - referred to as the \emph{universe shape} - via a one-to-one point-wise correspondence. Pairwise correspondences are then constructed by composing the shape-to-universe and universe-to-shape maps, inherently ensuring global consistency \cite{huang2013consistent}. Specifically, let $\Pi_i$ denote the correspondence between shape $\mathcal{S}_i$ and the universe. The point-wise correspondence $\Pi_{ij}$ between $\mathcal{S}_i$ and $\mathcal{S}_j$ is given by:
\begin{equation}
    \Pi_{ij} = \Pi_i \Pi_j^{\top},
\label{universe_Pi}
\end{equation}
where first maps $\mathcal{S}_i$ to the universe and then back to $\mathcal{S}_j$.

Likewise, functional maps exhibit natural cycle consistency under the universe-based formulation \cite{sun2023spatially}, as formalized below.
\begin{theorem}
Let the total energy over all shape pairs be defined as $E_{\text{total}}(\mathcal{C}) = \sum_{i,j} E_{\text{data}} (C_{ij}) + \lambda E_{\text{reg}} (C_{ij})$. If $E_{\text{total}}(\mathcal{C}) = 0$, then for any shape $\mathcal{S}i$ and any closed cycle $\{i, j, \dots, l, i\}$, the composed functional map satisfies $C_{ii} \mathcal{A}_i = \mathcal{A}_i$, i.e., it acts as the identity on the subspace spanned by $\mathcal{A}_i$.
\label{the 1}
\end{theorem}

In practice, $\mathcal{A}_i$ is typically of full row rank, as we generally set $d > k_{\mathrm{LB}} + k_{\mathrm{Elas}}$. Under this condition, Theorem \ref{the 1} implies $C_{ii} = \mathbf{I}$, from which it follows that the entire set $\mathcal{C} = \{C_{ij}\}_{i,j=1}^{n}$ is cycle consistent.
In this case, $\mathcal{A}_i$ can be interpreted as a functional embedding of shape $\mathcal{S}_i$ into a shared universe. Analogous to Eq.~\eqref{universe_Pi}, the functional map between any two shapes can be expressed as:
\begin{equation}
    C_{ij} = \mathcal{A}_j \mathcal{A}_i^{\dagger}.
\end{equation}

\section{Method}
In this section, we detail our proposed method, as illustrated in Fig.~\ref{fig:framework}. The pipeline begins with feature extraction, followed by the functional maps component. We then introduce the shape graph attention module, which captures inter-shape relationships across the shape set. Subsequently, the shape-to-universe correspondences are established through the universe predictor module. Finally, we describe the unsupervised loss functions used to train the network.

\subsection{Feature Extractor}
We begin by extracting features from a set of input shapes $\mathcal{S} = \{ \mathcal{S}_i\}_{i=1}^{n}$. For each shape, we compute per-vertex features that serve as the foundation for all subsequent components. We employ DiffusionNet \cite{sharp2022diffusionnet} as our feature extractor, owing to its robustness to variations in mesh resolution and sampling density. This ensures consistent and discriminative features across diverse shape collections. The extracted features are denoted as $\mathcal{F} = \{ \mathcal{F}_i\}_{i=1}^{n}$.

\subsection{Functional Maps Module}
For notational simplicity, we consider a shape pair $\mathcal{S}_i$ (source) and $\mathcal{S}_j$ (target) from the collection $\mathcal{S} = \{ \mathcal{S}_i\}_{i=1}^{n}$ when describing operations between two shapes. 

The functional maps module is responsible for establishing direct correspondences between shape pairs. Specifically, it computes bidirectional functional maps $C_{ij}$ and $C_{ji}$, along with the corresponding point-wise map $\Pi_{ij}$ between $\mathcal{S}_i$ and $\mathcal{S}_j$. This module comprises two components: a functional map solver and a point-wise map solver, both of which take as input the features produced by the feature extractor.

\subsubsection{Functional Maps Solver}
We employ a regularized functional map formulation to estimate $C_{ij}$ and $C_{ji}$. During training, we incorporate regularization terms promoting bijectivity and orthogonality, following \cite{ren2019structured}:
\begin{equation}
    \mathcal{L}_{\text{struct}} = \lambda_{\text{bij}} \mathcal{L}_{\text{bij}} + \lambda_{\text{orth}} \mathcal{L}_{\text{orth}},
\label{eq:struct_loss}
\end{equation}
where
\begin{equation}
\textstyle{
    \mathcal{L}_{\text{bij}} = \sum_{i,j}^{n} \left\| C_{ij} C_{ji} - \mathbf{I} \right\|_\mathrm{F}^2 + \left\| C_{ji} C_{ij} - \mathbf{I} \right\|_\mathrm{F}^2,}
\end{equation}
\begin{equation}
\textstyle{
    \mathcal{L}_{\text{orth}} = \sum_{i,j}^{n} \left\| C_{ij}^* C_{ji} - \mathbf{I} \right\|_\mathrm{F}^2 + \left\| C_{ji}^* C_{ij} - \mathbf{I} \right\|_{\mathrm{F}}^2.}
\end{equation}

\subsubsection{Point-wise Map Solver}
The point-wise map $\Pi_{ij}$ is theoretically expected to be a (partial) permutation matrix:
\begin{equation}
    \left\{ \Pi_{ij} \in \{0, 1\}^{n_i \times n_j}: \Pi_{ij} \textbf{1}_{n_j} = \textbf{1}_{n_i}, \, \textbf{1}_{n_i}^\top \Pi_{ij} \leq \textbf{1}_{n_j}^\top \right\},
\end{equation}
where $\Pi_{ij}(s,t)$ indicates the correspondence between the $s$-th vertex of $\mathcal{S}_i$ and the $t$-th vertex of $\mathcal{S}_j$. Following \cite{eisenberger2021neuromorph}, we estimate $\Pi_{ij}$ by computing the similarity between feature matrices $\mathcal{F}_i$ and $\mathcal{F}_j$:
\begin{equation}
    \Pi_{ij} = \text{softmax}(\mathcal{F}_i \mathcal{F}_j^{\top} / \tau),
\label{eq:compute Pi}
\end{equation}
where $\tau$ is a temperature parameter that controls the sharpness of the assignment distribution.

\subsection{Shape Graph Attention Module}
Shape collections typically exhibit structured relationships rather than consisting of independent entities. Some shapes are more similar to each other, and mappings between similar shapes often share intrinsic geometric correlations. To capture these dependencies, we first construct a shape graph and employ a Graph Attention Network (GAT) \cite{brody2022how} to refine the per-shape features $\mathcal{F} = \{ \mathcal{F}_i \}_{i=1}^{n}$. The edge set $\mathcal{E}$ of the graph is defined based on the top-$k$ cosine similarities between shape features:
\begin{equation}
\mathcal{E} = \left\{ (i, j) ,\middle|, j \in \text{Top-}k\big(\operatorname{cos}(\mathcal{F}_i, \mathcal{F}_j)\big) \right\}.
\label{eq:top-k}
\end{equation}
This connects each shape to its most similar neighbors, forming a graph that encodes the underlying geometric structure of the shape set.

To extract manifold-aware features, we employ a GAT, which dynamically learns attention weights between connected shapes. Unlike fixed edge weights, the attention mechanism adaptively computes inter-shape affinities, enabling more flexible and expressive feature aggregation. We begin by applying mean pooling to the per-vertex features of each shape, resulting in a shape-level descriptor. Given a pair of shapes $(i, j)$, the attention weight $\alpha_{ij}$ is computed as:
\begin{equation}
    \alpha_{ij} = \frac{ \exp \left( \mathbf{a}^\top \text{LeakyReLU} \left( \mathbf{W} \cdot [\mathcal{F}_i \| \mathcal{F}_j] \right) \right) }{ \sum_{j' \in \mathcal{N}_i} \exp \left( \mathbf{a}^\top \text{LeakyReLU} \left( \mathbf{W} \cdot [\mathcal{F}_i \| \mathcal{F}_j'] \right) \right) },
\end{equation}
where $\mathbf{W}$ is a learnable weight matrix and $\mathbf{a}$ is a learnable attention vector. The final manifold-aware feature for shape $\mathcal{S}_i$ is obtained by aggregating features from its neighbors, weighted by attention:
\begin{equation}
\textstyle{
    \mathcal{F}_i^{'} = \sigma \left( \sum_{j \in \mathcal{N}_i} \alpha_{ij} \cdot \mathbf{W} \mathcal{F}_j \right),}
\end{equation}
where $\sigma$ denotes a non-linear activation function. We use a two-layer GAT to model the local structure of the shape graph, followed by a LayerNorm and dropout on the final output to improve stability and generalization.

Through message passing on the shape graph, each shape's representation is enriched with contextual information from its neighbors, capturing the underlying manifold structure. To preserve both local geometric detail and global structural context, we concatenate the original features $\mathcal{F}_i$ with the aggregated features $\mathcal{F}_i^{'}$ to form the final representation $\mathcal{G} = \{ \mathcal{G}_i \}_{i=1}^n$, where $\mathcal{G}_i = [\mathcal{F}_i^{'} \| \mathcal{F}_i]$.

\begin{table*}[t]
    \centering
    \small
    \caption{\textbf{Quantitative results on near-isometric datasets (FAUST, SCAPE, SHREC’19) and anisotropically remeshed versions (FAUST\_a, SCAPE\_a).} The best results are shown in \textbf{bold}.}
    \renewcommand{\arraystretch}{1.1}
    \setlength{\tabcolsep}{8pt}

    \begin{tabular}{cl cc cc cccc}
    \toprule
        & Train & \multicolumn{2}{c}{\textbf{FAUST}} &  \multicolumn{2}{c}{\textbf{SCAPE}} & \multicolumn{3}{c}{\textbf{FAUST+SCAPE}} \\
        \cmidrule(lr){3-4}  \cmidrule(lr){5-6} \cmidrule(lr){7-9}
        & Test  & \textbf{FAUST} & \textbf{FAUST\_a} & \textbf{SCAPE} & \textbf{SCAPE\_a} & \textbf{FAUST} & \textbf{SCAPE} & \textbf{SHREC'19} \\
	\midrule
	\multirow{9}{*}{\rotatebox{90}{\emph{Pairwise Matching}}}
	& ZoomOut      & 6.1 & 8.7 & 7.5  & 14.0 & 6.1 & 7.5  & - \\
	& SmoothShells & 2.5 & 5.4 & 4.7  & 5.0  & 2.5 & 4.7  & - \\
	& DiscreteOp   & 5.6 & 6.2 & 13.1 & 14.6 & 5.6 & 13.1 & - \\
	
	& Deep Shells    & 1.7 & 12.0 & 2.5 & 10.0 & 1.6 & 2.4 & 21.1 \\
	& DUO-FMNet      & 2.5 & 3.0  & 2.6 & 2.7  & 2.5 & 4.3 & 6.4 \\
	& AttentiveFMaps & 1.9 & 2.4  & 2.2 & 2.3  & 1.9 & 2.3 & 5.8 \\
	& ULRSSM         & 1.6 & 1.9  & 1.9 & 1.9  & 1.6 & 2.1 & 4.8 \\
	& HybridFMaps    & 1.5 & 1.8  & \textbf{1.8} & 1.9  & 1.5 & 2.0 & 4.5 \\
	& DenoisFMaps    & 1.7 & 2.0  & 2.1 & 2.2  & 1.7 & 2.0 & 6.3 \\
	
	\midrule
	\multirow{6}{*}{\rotatebox{90}{\emph{Multi-Matching}}}
	& Consistent ZoomOut  & 2.2 & 7.3  & 2.5 & 12.1 & 2.2 & 2.5 & - \\ 
	& IsoMuSh             & 4.4 & -    & 5.6 & -    & 4.4 & 5.6 & - \\
	& CycoMatch           & 3.8 & 9.2  & 4.7 & 13.6 & 3.8 & 4.6 & - \\
	& UDMSM               & 1.5 & 15.3 & 2.0 & 4.9  & 1.7 & 3.2 & 17.8 \\
	& G-MSM               & 1.5 & 12.7 & \textbf{1.8} & 28.1 & 1.5 & 2.1 & 6.8 \\
	& \textbf{Ours}     & \textbf{1.4} & \textbf{1.7} & \textbf{1.8} & \textbf{1.8} & \textbf{1.4} & \textbf{1.9} & \textbf{4.2}\\
    \bottomrule
    \end{tabular}
    \label{tab:near_anisotropic}
\end{table*}

\subsection{Universe Predictor Module}
The universe predictor takes as input the shape-level features $\mathcal{G} = \{ \mathcal{G}_i \}_{i=1}^n$ produced by the graph attention module, and predicts a correspondence $\Pi_i$ that maps each shape $\mathcal{S}_i$ to the shared universe. Following UDMSM \cite{cao2022unsupervised}, we utilize a DiffusionNet architecture to generate these assignment matrices, where the number of output columns corresponds to the number of universe points.

In the universe-based formulation, each vertex in a shape is assigned to a single universe point, while each universe point maps to at most one shape vertex. Thus, $\Pi_i$ is a (partial) permutation matrix:
\begin{equation}
    \Pi_i \in \left\{ \Pi \in \{0, 1\}^{n_i \times c}: \Pi \textbf{1}_c = \textbf{1}_{n_i}, \, \textbf{1}_{n_i}^\top \Pi \leq \textbf{1}_c^\top \right\},
\end{equation}
where $c$ denotes the number of universe points.

To enable end-to-end training, we apply Sinkhorn normalization \cite{sinkhorn1967concerning, mena2018learning} to the raw network outputs. This iterative normalization projects assignment scores into a doubly stochastic matrix, serving as a smooth, differentiable approximation of a partial permutation. It enables the network to learn discrete constraints in a relaxed yet consistent manner, supporting optimization in a fully unsupervised setting.

\subsection{Loss Functions}
\subsubsection{Spectral Loss}
The spectral loss integrates both structural regularization (Eq.~\eqref{eq:struct_loss}) and a coupling term. The coupling loss $\mathcal{L}_{\text{couple}}$ encourages the point-wise maps $\Pi_{ij}$ and $\Pi_{ji}$ to be consistent via the functional maps:
\begin{equation}
\textstyle{
    \mathcal{L}_{\text{couple}} = \sum_{i,j}^{n} \left\| C_{ij} - \Phi_j^{\dagger} \Pi_{ji} \Phi_i \right\|_F^2 + \left\| C_{ij} - \Psi_i^{\dagger} \Pi_{ji} \Psi_j \right\|_{\mathrm{HS}}^2.}
\end{equation}
The total spectral loss is given by:
\begin{equation}
\mathcal{L}_{\text{spectral}} = \mathcal{L}_{\text{struct}} + \lambda_{\text{couple}} \mathcal{L}_{\text{couple}}.
\label{eq:spectral_loss}
\end{equation}

\subsubsection{Cycle Consistency Loss}
In our framework, shape-to-universe alignment can be realized in two compatible forms. The functional map coefficient matrix $\mathcal{A}_i$ maps the spectral basis of shape $\mathcal{S}_i$ to a shared latent universe, resulting in the aligned embedding $\Phi_i \mathcal{A}_i$. Alternatively, the point-wise correspondence matrix $\Pi_i$ projects shape vertices directly onto the universe, yielding a universe-aligned embedding $\Pi_i^{\top} \Phi_i$.

To exploit this structural redundancy, we introduce a novel cycle consistency loss that enforces consistency between these two alignment paths across the shape collection. This constraint leverages the intrinsic cycle consistency of functional maps while regularizing the learning of shape-to-universe matching matrices, enhancing robustness against noise and structural variability.

We consider two variants of this loss. For near-isometric shapes, we use a Frobenius norm formulation:
\begin{equation}
\textstyle{
    \mathcal{L}_{\text{cycle}} = \sum_{i,j}^{n} \left\| \Pi_i^{\top} \widetilde{\Phi}_i \mathcal{A}_i - \Pi_j^{\top} \widetilde{\Phi}_j \mathcal{A}_j \right\|_F^2.}
\end{equation}

For more general deformations, we adopt a cosine similarity formulation:
\begin{equation}
\textstyle{
    \mathcal{L}_{\text{cycle}} = \sum_{i,j}^{n} \left( 1 - \text{cos}\left( \Pi_i^{\top} \widetilde{\Phi}_i \mathcal{A}_i, \, \Pi_j^{\top} \widetilde{\Phi}_j \mathcal{A}_j \right) \right).}
\end{equation}

The final loss combines the spectral and cycle terms:
\begin{equation}
\mathcal{L}_{\text{total}} = \mathcal{L}_{\text{spectral}} + \lambda_{\text{cycle}} \mathcal{L}_{\text{cycle}}.
\label{eq:total_loss}
\end{equation}

\begin{figure*}[t]
\centering
\begin{subfigure}{0.32\textwidth}
    \centering
    \includegraphics[width=\textwidth]{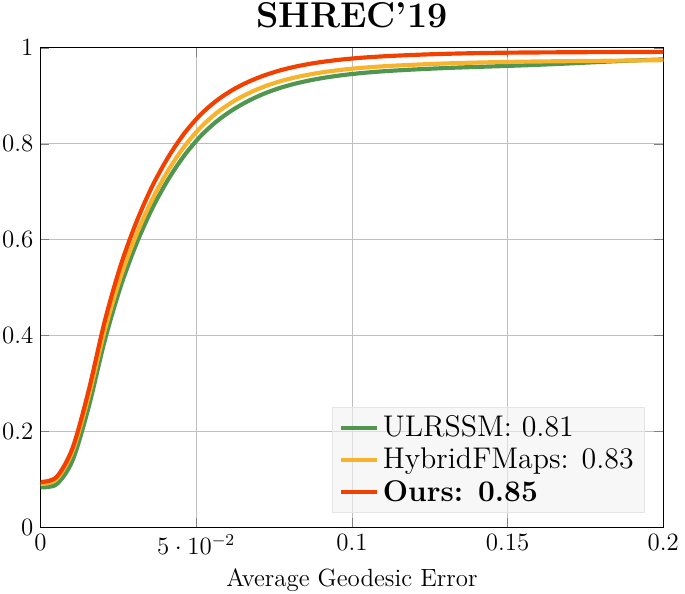}
\end{subfigure}
\hfill
\begin{subfigure}{0.32\textwidth}
    \centering
    \includegraphics[width=\textwidth]{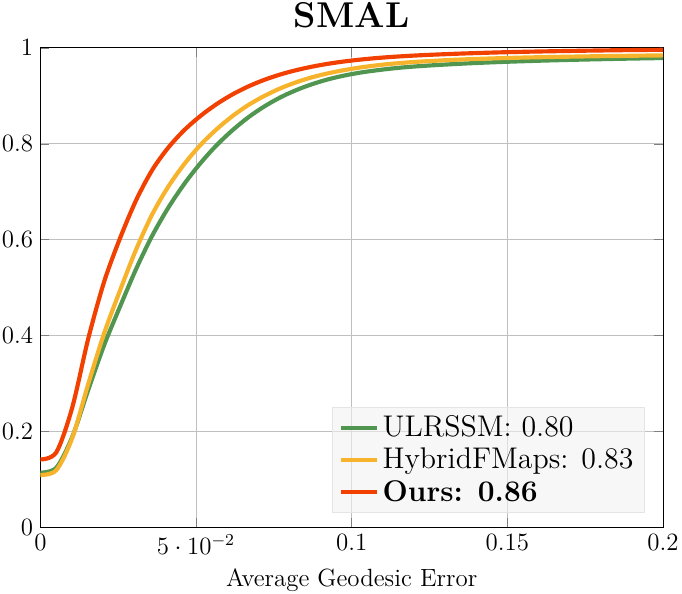}
\end{subfigure}
\hfill
\begin{subfigure}{0.32\textwidth}
    \centering
    \includegraphics[width=\textwidth]{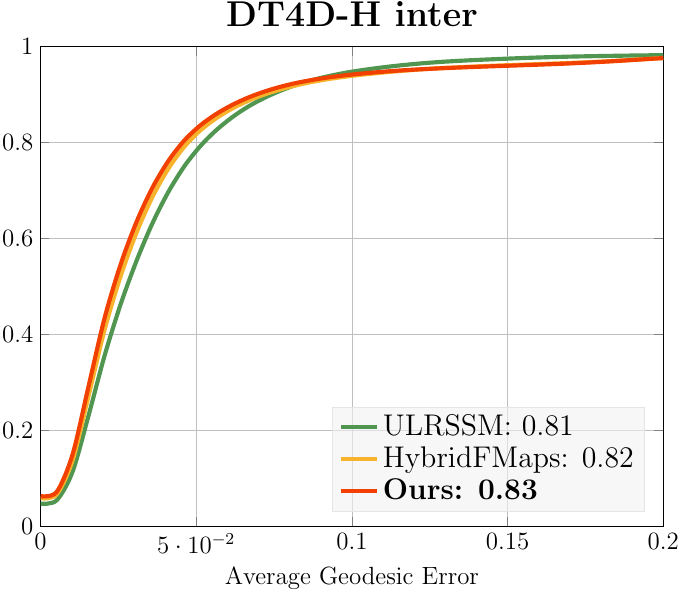}
\end{subfigure}
\caption{\textbf{Proportion of Correct Keypoints (PCK) curves and Area Under Curve (AUC) values} on SHREC’19, SMAL, and DT4D-H inter-class, comparing our method with ULRSSM and HybridFMaps.}
\label{fig:pcks}
\end{figure*}

\begin{figure*}[t]
\centering
\includegraphics[width=0.94\textwidth]{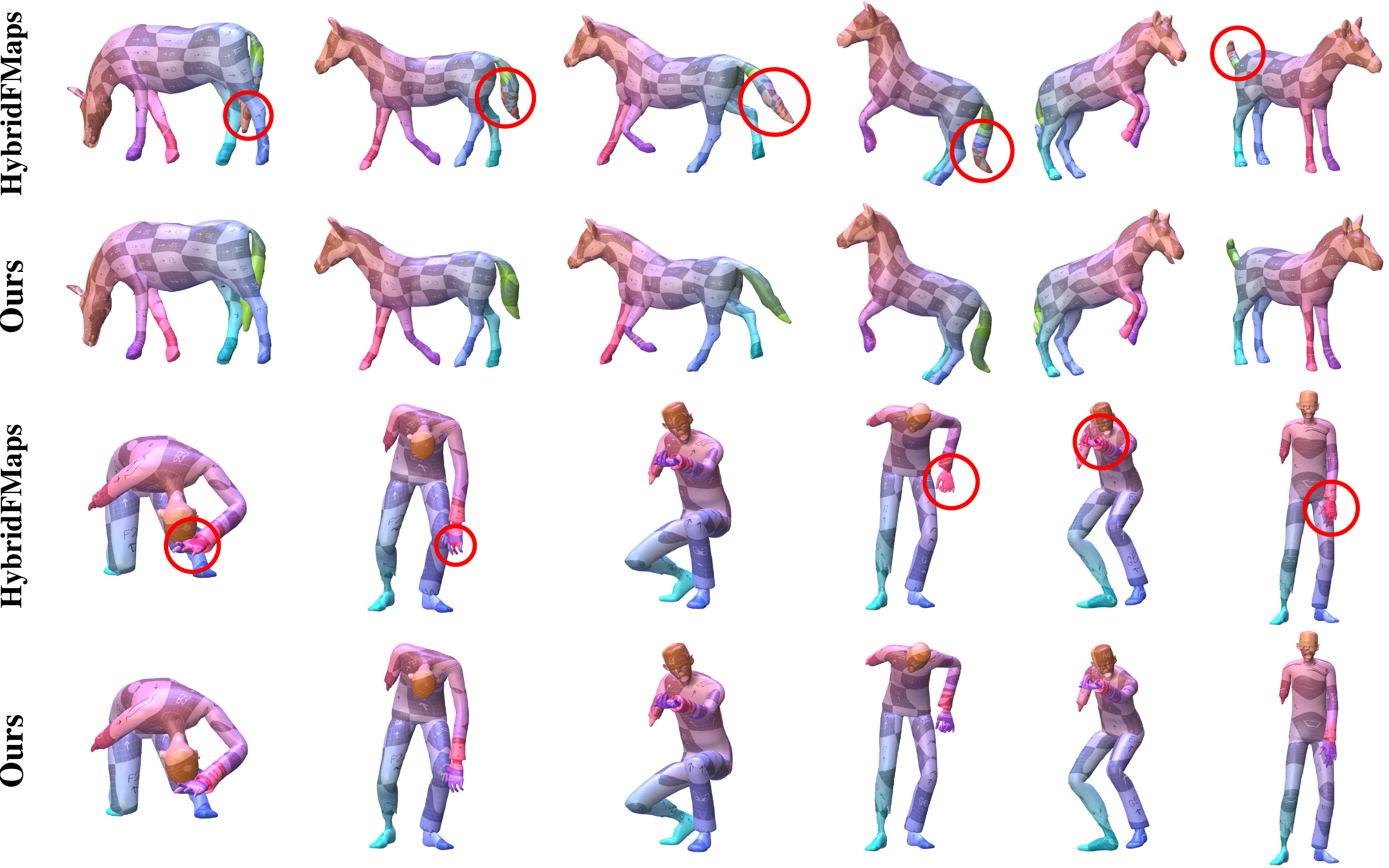}
\caption{\textbf{Qualitative multi-shape matching results} via texture transfer, comparing our method with HybridFMaps on SMAL (top) and DT4D-H inter-class (bottom).}
\label{fig:comparsion}
\vspace{-0.15in}
\end{figure*}

\section{Experiment}
In this section, we evaluate our approach on multiple benchmarks and compare it against state-of-the-art methods.

\paragraph{Baselines} We compare our method with a wide range of baselines, including pairwise methods (ZoomOut \cite{melzi2019zoomout}, SmoothShells \cite{eisenberger2020smooth}, DiscreteOp \cite{ren2021discrete}, DeepShells \cite{eisenberger2020deep}, DUO-FMNet \cite{donati2022deep}, AttentiveFMaps \cite{li2022learning}, ULRSSM \cite{cao2023unsupervised}, HybridFMaps \cite{bastian2024hybrid}, DenoisFMaps \cite{zhuravlev2025denoisfm}) and multi-matching approaches (ConsistentZoomOut \cite{huang2020consistent}, IsoMuSh \cite{gao2021isometric}, CycoMatch \cite{xia2025multi}, UDMSM \cite{cao2022unsupervised}, G-MSM \cite{eisenberger2023g}).

\paragraph{Evaluation Metrics} Following the Princeton benchmark protocol \cite{kim2011blended}, we report the average geodesic error (×100) as the evaluation metric.

\subsection{Near-isometric Shape Matching}
\subsubsection{Datasets}
We evaluate on three near-isometric datasets: FAUST \cite{bogo2014faust}, SCAPE \cite{anguelov2005scape}, and SHREC’19 \cite{melzi2019shrec}, using their remeshed versions \cite{ren2018continuous, donati2020deep}. FAUST contains 100 human shapes across 10 subjects and 10 poses, with a 80/20 train-test split. SCAPE includes 71 shapes from a single subject, split 51/20 for training and testing. SHREC’19 consists of 44 human shapes with more variation in body type and pose, used solely for testing, excluding shape 40 due to its non-closed geometry.

\subsubsection{Results}
We compare our method with both pairwise shape matching algorithms and multi-shape matching frameworks. Table \ref{tab:near_anisotropic} reports the average geodesic error for all methods. Compared to pairwise approaches, our method consistently achieves superior performance across most settings and datasets. When compared to multi-shape baselines, it demonstrates strong generalization, particularly on the SHREC’19 dataset. We attribute this to the model’s ability to capture the underlying manifold structure of the shape collection, leading to more robust correspondences on unseen shapes. The Proportion of Correct Keypoints (PCK) curves on the SHREC'19 dataset are shown in Fig.~\ref{fig:pcks} (left).

\subsection{Matching with Anisotropic Meshing}
\subsubsection{Datasets}
To assess robustness to mesh connectivity variations, we follow DUO-FMNet \cite{donati2022deep} and use anisotropically remeshed versions of FAUST and SCAPE (FAUST\_a and SCAPE\_a), where feature uneven triangulations with small, dense triangles on one side and large, coarse triangles on the other. These distortions challenge methods sensitive to mesh structure.

\subsubsection{Results}
Table~\ref{tab:near_anisotropic} presents the quantitative results. Most methods exhibit a noticeable drop in performance under anisotropic conditions. In contrast, our method achieves the best performance on both datasets. These results suggest that the combination of pairwise functional maps and manifold-aware universe matching enables our approach to effectively mitigate the challenges posed by irregular mesh structures.

\begin{table}[t]
\centering
\small
\caption{\textbf{Quantitative results on non-isometric shape matching.}}
\renewcommand{\arraystretch}{1.1}
\setlength{\tabcolsep}{8pt}
\begin{tabular}{cl ccc}
    \toprule
    & \multirow{2}{*}{\textbf{Geo.error ($\times 100$)}} & \multirow{2}{*}{\textbf{SMAL}} & \multicolumn{2}{c}{\textbf{DT4D-H}} \\
    \cmidrule(lr){4-5}
    & & &  \textbf{intra} & \textbf{inter} \\
    \midrule

    \multirow{9}{*}{\rotatebox{90}{\emph{Pairwise Matching}}}
    & ZoomOut & 38.4 & 4.0 & 29.0 \\
    & SmoothShells  & 36.1 & 1.1 & 6.3  \\
    & DiscreteOp  & 38.1 & 3.6 & 27.6  \\
    
    & Deep Shells  & 29.3 & 3.4 & 31.1  \\
    & DUO-FMNet & 6.7 & 2.6 & 15.8   \\
    & AttentiveFMaps  & 5.4 & 1.7 & 11.6  \\
    & ULRSSM  & 3.9 & \textbf{0.9} & 4.1  \\
    & HybridFMaps  & 3.4 & 1.0 & 3.9 \\
    & DenoisFMaps  & 46.1 &  14.4 & 22.9 \\
    
    \midrule
    \multirow{5}{*}{\rotatebox{90}{\small \emph{Multi-Matching}}}
    & Consistent ZoomOut  & 16.9 & 8.7 & 26.4 \\
    & CycoMatch  & 24.7 & 4.4  & 24.9 \\
    
    & UDMSM  & 26.5 & 2.4 & 15.8 \\
    & G-MSM  & 43.9 & 7.8 & 12.0 \\
    & \textbf{Ours} & \textbf{2.9} & 1.0 & \textbf{3.8} \\
    \bottomrule
\end{tabular}
\label{tab:non_isometric}
\end{table}

\subsection{Non-isometric Shape Matching}
\subsubsection{Datasets}
We also evaluate on two widely used non-isometric datasets: SMAL \cite{zuffi20173d} and DT4D-H \cite{magnet2022smooth}. The SMAL dataset contains 49 animal shapes from 8 species; we use 5 species for training and the remaining for testing, yielding a 29/20 train-test split. The DT4D-H dataset includes 9 human shape categories, with 198 shapes used for training and 95 for testing.

\subsubsection{Results}

Table~\ref{tab:non_isometric} presents results for both intra-class and inter-class settings. The DT4D-inter setup is particularly challenging, as it requires generalization to unseen shape classes. Our method achieves competitive intra-class results and significantly outperforms all baselines in the inter-class case, with notable improvements on the SMAL. These results demonstrate the strong generalization of our approach across diverse, non-isometric categories. PCK curves are shown in Fig.~\ref{fig:pcks}, with qualitative comparisons in Fig.~\ref{fig:comparsion}.

\begin{table}[t]
\centering
\caption{\textbf{Ablation study results on the SMAL dataset.}}
\renewcommand{\arraystretch}{1.1}
\setlength{\tabcolsep}{8pt}
\begin{tabular}{lc}
    \toprule
    \textbf{Ablation Setting} & \textbf{Geo.error ($\times 100$)}\\
    \midrule
    w/o shape graph attention module & 3.7   \\
    w/o functional maps module & 26.5 \\
    w/o universe predictor module & 3.4 \\
    w/o cycle consistency loss & 3.8\\
    \textbf{Ours} & \textbf{2.9} \\
    \bottomrule
\end{tabular}
\label{tab:ablation}
\end{table}

\subsection{Ablation Study}
We conduct ablation experiments on the non-isometric SMAL dataset to evaluate the contribution of each component in our framework. Specifically, we compare the following configurations: (1) Removing the shape graph attention module; (2) Removing the functional map module; (3) Removing the universe predictor module; (4) Removing the cycle consistency loss. The results in Table~\ref{tab:ablation} show that removing any single component leads to a noticeable performance drop, especially the functional maps module.

\section{Limitations}
Despite its effectiveness, our method has several limitations that merit further exploration. First, although our framework avoids the need to explicitly select a reference shape, it still requires the universe size to be predefined and fixed. Second, processing the entire shape collection as a graph incurs additional computational overhead. While this trade-off leads to improved matching accuracy, future work could focus on designing more efficient and scalable algorithmic alternatives to reduce runtime and memory demands.

\section{Conclusion}
In this paper, we introduce DcMatch, an unsupervised multi-shape matching framework with dual-level cycle consistency. Using a shape graph attention network, our method captures the manifold structure of shape collections, enabling more informed and robust feature aggregation. By enforcing spatial and spectral cycle consistency, DcMatch improves accuracy and coherence in the shared universe space. Extensive experiments across multiple benchmarks and ablation studies validate the superior performance and effectiveness of its components. We believe this work advances multi-shape matching by highlighting the importance of manifold-aware modeling and dual-level consistency.

\appendix
\section{Acknowledgments}
This work was supported by the National Key R\&D Program of China (No.2024YFE0202500), the Natural Science Foundation of Guangdong Province, China (2023A1515012834), the Natural Science Foundation of Hubei Province, China (2025AFB639) and the Natural Science Foundation of China (U23B2050).

\bibliography{ref}

\newpage
In this supplementary material, we first provide the proof of Theorem 1, followed by additional implementation details. We then present runtime and memory analysis and include additional ablation experiment results. Finally, we further provide statistical stability analysis and qualitative examples to complement the visualizations.

\section{Proof of Theorem 1}
\begin{theorem}
Let the total energy over all shape pairs be defined as $E_{\text{total}}(\mathcal{C}) = \sum_{i,j} E_{\text{data}} (C_{ij}) + \lambda E_{\text{reg}} (C_{ij})$. If $E_{\text{total}}(\mathcal{C}) = 0$, then for any shape $\mathcal{S}i$ and any closed cycle $\{i, j, \dots, l, i\}$, the composed functional map satisfies $C_{ii} \mathcal{A}_i = \mathcal{A}_i$, i.e., it acts as the identity on the subspace spanned by $\mathcal{A}_i$.
\end{theorem}

\begin{proof}
If $E_{\text{total}}(\mathcal{C}) = 0$, then in particular the data term vanishes, i.e., $E_{\text{data}}(C_{ij}) = 0$ for all $(i,j)$. This implies:
\begin{equation}
C_{ij} \mathcal{A}_i = \mathcal{A}_j.
\end{equation}
Now consider a simple cycle involving three shapes, e.g., $(i, j, k, i)$. The composed functional map along this cycle is given by $C_{ii} = C_{ki} C_{jk} C_{ij}$. Applying this to $\mathcal{A}_i$ yields:
\begin{equation}
\begin{aligned}
C_{ii} \mathcal{A}_i &= C_{ki} C_{jk} C_{ij} \mathcal{A}_i 
= C_{ki} C_{jk} \mathcal{A}_j 
= C_{ki} \mathcal{A}_k 
= \mathcal{A}_i.
\end{aligned}
\end{equation}
Hence, $C_{ii} \mathcal{A}_i = \mathcal{A}_i$, as required.
\end{proof}

\section{Implementation Details}
All learning-based methods are implemented in PyTorch 2.1.0 with CUDA 12.1. Axiomatic baselines such as ZoomOut, SmoothShells, IsoMuSh, and CycoMatch are implemented in MATLAB R2018a, using the official code provided by the authors. All experiments are run on an NVIDIA TITAN RTX GPU and an Intel i9-9920X CPU (3.50 GHz).

We use DiffusionNet as the feature extractor with WKS descriptors as input. For the SMAL dataset, we instead use raw XYZ coordinates and apply random rotation augmentations. The output feature dimension is fixed to 256 across all experiments.

Training is performed end-to-end using the Adam optimizer with a learning rate of 0.001. 
The functional maps module uses $\lambda_{\mathrm{LB}} = 100$, $\lambda_{\mathrm{Elas}} = 50$, and a temperature parameter $\tau = 0.07$. We set the number of basis functions to $k_{\mathrm{LB}} = 160$ and $k_{\mathrm{Elas}} = 40$. Loss weights are set as $\lambda_{\text{bij}} = \lambda_{\text{orth}} = \lambda_{\text{couple}} = \lambda_{\text{cycle}} = 1.0$.

During inference, point-wise correspondences are computed as $\Pi_{ij} = \Pi_i \Pi_j^{\top}$.

For the selection of universe size $c$, if ground-truth correspondences are defined relative to a reference shape, we set the number of universe points to match its vertex count. Otherwise, we use the maximum vertex count across the dataset.

For the parameter $k$ in Eq.\eqref{eq:top-k}, we empirically set it to 3, given that each input consists of 5 shapes. This choice strikes a balance between maintaining local geometric relationships and capturing broader contextual information.

\begin{table}[!ht]
\renewcommand \arraystretch{1.1}
\centering
\small
\caption{Summary of notations used in this paper.}
\begin{tabular}{ll}
    \toprule
    \textbf{Notation} & \textbf{Description} \\
    \midrule
    $\mathcal{S} = \{ \mathcal{S}_i \}_{i=1}^n$ & Set of $n$ input shapes \\
    $\mathcal{P} = \{ \Pi_{ij} \}$ & Point-wise maps between shape pairs \\
    $\mathcal{C} = \{ C_{ij} \}$ & Functional maps between shape pairs \\
    $\Phi_i \in \mathbb{R}^{n_i \times k_{\mathrm{LB}}}$ & LBO basis on shape $\mathcal{S}_i$ \\
    $\Psi_i \in \mathbb{R}^{n_i \times k_{\mathrm{Elas}}}$ & Elastic basis on shape $\mathcal{S}_i$ \\
    $\mathcal{A}_i^{\mathrm{LB}} \in \mathbb{R}^{k_{\mathrm{LB}} \times d}$ & LBO spectral coefficients \\
    $\mathcal{A}_i^{\mathrm{Elas}} \in \mathbb{R}^{k_{\mathrm{Elas}} \times d}$ & Elastic spectral coefficients \\
    $\mathcal{A}_i \in \mathbb{R}^{(k_{\mathrm{LB}} + k_{\mathrm{Elas}}) \times d}$ & Hybrid spectral features \\
    $\widetilde{\Phi}_i \in \mathbb{R}^{n_i \times (k_{\mathrm{LB}} + k_{\mathrm{Elas}})}$ & Hybrid basis (LBO + elastic) \\
    $\mathcal{F} = \{ \mathcal{F}_i \}$ & Per-vertex features \\
    $C_{ij} \in \mathbb{R}^{(k_{\mathrm{LB}} + k_{\mathrm{Elas}})^2}$ & Functional map from $\mathcal{S}_i$ to $\mathcal{S}_j$ \\
    $\Pi_{ji} \in \mathbb{R}^{n_j \times n_i}$ & Point-wise map from $\mathcal{S}_j$ to $\mathcal{S}_i$ \\
    $\Pi_i \in \mathbb{R}^{n_i \times c}$ & Point-wise map from $\mathcal{S}_i$ to universe \\
    \bottomrule
\end{tabular}
\end{table}

\section{Runtime and Memory Analysis}
We evaluate the average per-pair inference time (in seconds) and peak GPU memory usage (in GB) of our method, comparing it with ULRSSM and HybridFMaps across several representative datasets. As shown in Table~\ref{tab:runtime}, our model consumes more memory than the baselines. This increase in memory usage is a result of the added model complexity, which contributes to improved matching consistency and accuracy. However, we recognize the potential for more lightweight designs and plan to explore this direction in future work. Regarding runtime, our method achieves comparable inference times to the baselines on near-isometric datasets (FAUST, SCAPE, SHREC'19). On more challenging datasets with larger non-isometric deformations (e.g., SMAL, DT4D), our method incurs a moderate increase in runtime, yet it remains within a practical and acceptable range for real-world applications.

\begin{table*}[!ht]
\centering
\caption{Runtime and memory analysis. We compare the average per-pair inference time (s) and peak GPU memory (GB) with ULRSSM and HybridFMaps.}
\renewcommand{\arraystretch}{1.1}
\begin{tabular}{lccccccc}
\toprule
\textbf{Method} & \textbf{Memory} & \textbf{FAUST} & \textbf{SCAPE} & \textbf{SHREC'19} & \textbf{SMAL} & \textbf{DT4D-H intra} & \textbf{DT4D-H inter} \\
\midrule
ULRSSM & 2.6 & 0.08 & 0.07 & 18.83 & 17.82 & 0.09 & 9.12 \\
HybridFMaps & 3.5 & 0.11 & 0.10 & 18.82 & 12.88 & 0.14 & 10.78 \\
\textbf{Ours} & 8.4 & 0.09 & 0.08 & 20.83 & 15.10 & 0.11 & 18.35 \\
\bottomrule
\end{tabular}
\label{tab:runtime}
\end{table*}

\section{Additional Ablation Experiment}
In this section, we present additional results and analysis to complement the ablation study. We first provide a more detailed discussion of the primary ablation, followed by further experiments investigating two specific design choices: (i) the number of universe points used in our model and (ii) different variants of the cycle consistency loss. 

\subsection{Detailed Analysis of Main Ablations}
As shown in Table~\ref{tab:ablation}, removing the functional maps module leads to the most significant performance drop, underscoring its critical role in establishing accurate correspondences. The shape graph attention module captures the manifold structure of the entire shape collection, enabling each shape to leverage contextual information from its neighbors. Without it, the universe embedding is learned from shapes in isolation, degrading the quality of shape-to-universe mappings. Excluding the universe predictor reduces our framework to a purely pairwise matching approach, relying almost entirely on functional maps for correspondence quality. Finally, removing the cycle consistency loss leaves the predictor’s shape-to-universe mappings unsupervised. Since these mappings are directly used for point-to-point correspondences during inference, this omission results in the second-largest performance degradation. These findings highlight the importance of enforcing alignment between spatial and spectral mappings in the shared universe space. Overall, the ablation results demonstrate that every module in our framework is essential for achieving accurate and consistent multi-shape matching.

\subsection{Effect of Universe Points}
We conduct an ablation study on the number of universe points using the SMAL dataset. As described earlier, we set the universe size to either the number of vertices in the reference shape or the maximum vertex count across the dataset, resulting in 5225 points for SMAL. Table~\ref{tab:ablation_universe} shows that a smaller universe size reduces memory usage but fails to capture the complexity of the shared embedding space, leading to suboptimal performance. Increasing the universe size further causes overfitting to the specific shape set, resulting in diminished accuracy and significantly higher inference time and memory consumption. These results validate our design choice, which provides a favorable balance between accuracy and computational efficiency.

\begin{table}[t]
\renewcommand \arraystretch{1.1}
\centering
\caption{Ablation Study on the number of universe points on the SMAL Dataset.}
\begin{tabular}{cccc}
    \toprule
    \textbf{Univ. Pts} & \textbf{\#Params} & \textbf{Infer. Time (s)} & \textbf{Geo. Err} \\
    \midrule
    1024 & 634k  & 14.1 & 3.8 \\
    2048 & 766k  & 29.2 & 3.4 \\
    4096 & 1.03M & 31.2 & 3.0\\
    5225 & 1.18M & 32.2 & 2.9 \\
    8192 & 1.56M & 35.1 & 3.2 \\
    \bottomrule
\end{tabular}
\label{tab:ablation_universe}
\end{table}

\subsection{Cycle Consistency Loss Variants}
We also examine the impact of different formulations of the cycle consistency loss. As shown in Table~\ref{tab:ablation_loss}, the Frobenius-based loss achieves comparable or superior performance on near-isometric datasets such as FAUST and SCAPE. In contrast, the cosine-based loss performs better on datasets with strong non-isometric deformations, such as SMAL and DT4D (inter-class). These results indicate that choosing an appropriate loss formulation according to the deformation characteristics of the data can lead to improved correspondence quality.

\begin{table}[t]
\renewcommand \arraystretch{1.1}
\centering
\caption{Ablation Study on the choice of cycle consistency loss variants.}
\begin{tabular}{cccc}
    \toprule
    \textbf{Dataset} & \textbf{Frobenius-based}  & \textbf{cosine-based} & \textbf{Ours} \\
    \midrule
    FAUST      & 1.4 & 1.5 & 1.4 \\
    SCAPE      & 1.8 & 2.0 & 1.8 \\
    SMAL       & 4.2 & 2.9 & 2.9 \\
    DT4D-H inter & 8.0 & 3.8 & 3.8 \\
    \bottomrule
\end{tabular}
\label{tab:ablation_loss}
\end{table}

\section{Statistical Stability Analysis}
We retrain and test our model on the SMAL dataset using five different random seeds (i.e., 42, 2023, 777, 999, 1234), reporting the mean and variance of the geodesic error. As shown in Table~\ref{tab:statis}, under identical conditions, our method achieves a lower mean geodesic error, with reduced variance, demonstrating superior and more stable performance.

\begin{table}[t]
\centering
\caption{Ablation results on the SMAL dataset: Geodesic error (mean ± standard deviation) for different methods.}
\renewcommand{\arraystretch}{1.1}
\setlength{\tabcolsep}{15pt}
\begin{tabular}{lc}
    \toprule
    \textbf{Methods} & \textbf{Geo.error (mean ± standard)}\\
    \midrule
    ULRSSM & 6.15 ± 1.92   \\
    HybridFMaps & 4.54 ± 1.26 \\
    \textbf{Ours} & \textbf{3.30 ± 0.98} \\
    \bottomrule
\end{tabular}
\label{tab:statis}
\end{table}

\section{Additional Qualitative Examples}
\begin{figure}[!ht]
\centering
\includegraphics[width=0.45\textwidth]{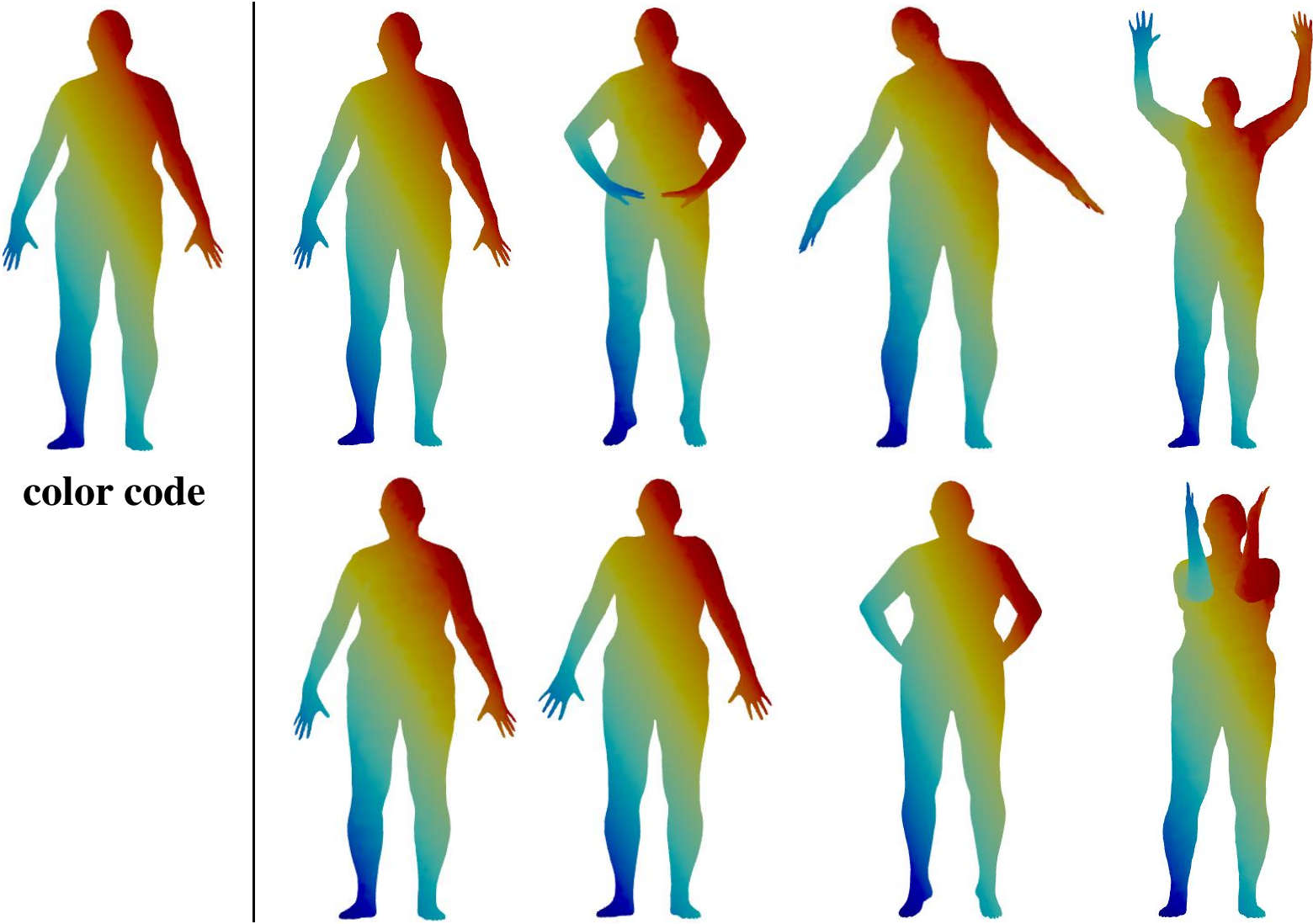}
\caption{\textbf{Qualitative examples on FAUST dataset.}}
\label{fig:comparsion_FAUST}
\end{figure}

\begin{figure}[!ht]
\centering
\includegraphics[width=0.5\textwidth]{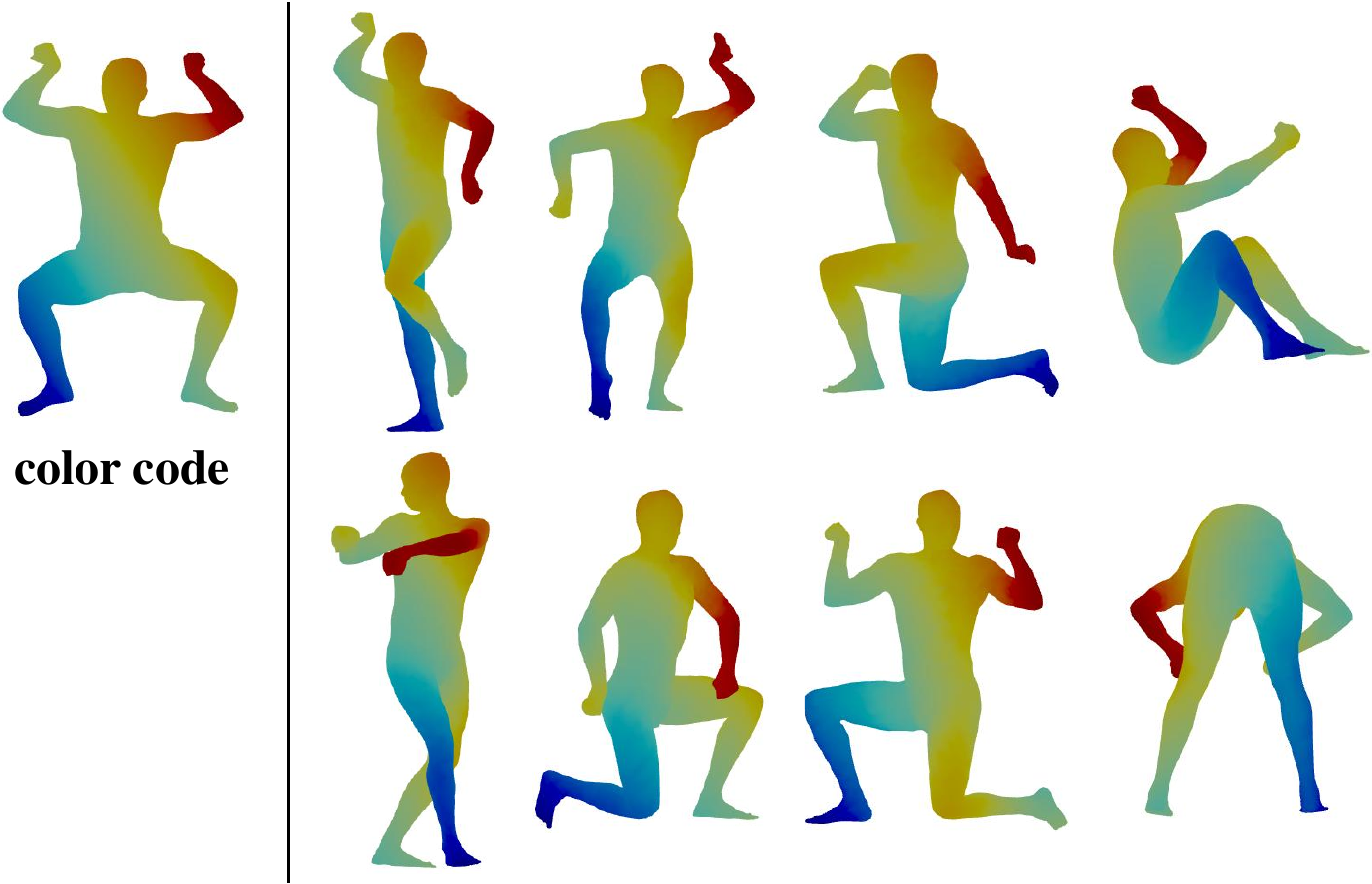}
\caption{\textbf{Qualitative examples on SCAPE dataset.}}
\label{fig:comparsion_SCAPE}
\end{figure}

\begin{figure}[!ht]
\centering
\includegraphics[width=0.5\textwidth]{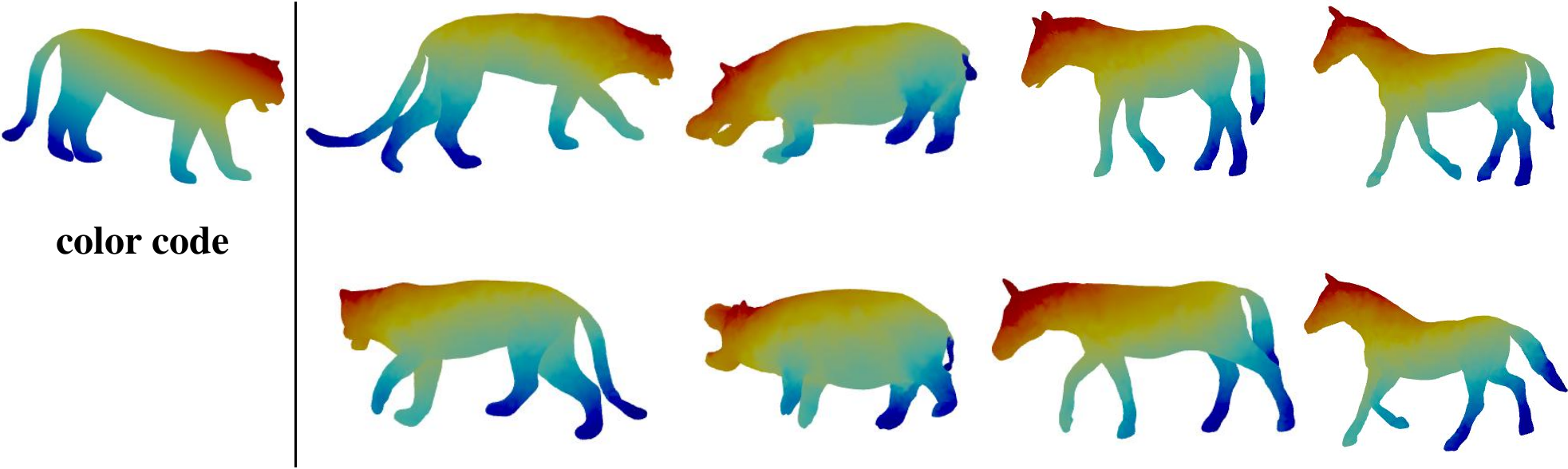}
\caption{\textbf{Qualitative examples on SMAL dataset.}}
\label{fig:comparsion_SMAL}
\end{figure}

\begin{figure}[!ht]
\centering
\includegraphics[width=0.5\textwidth]{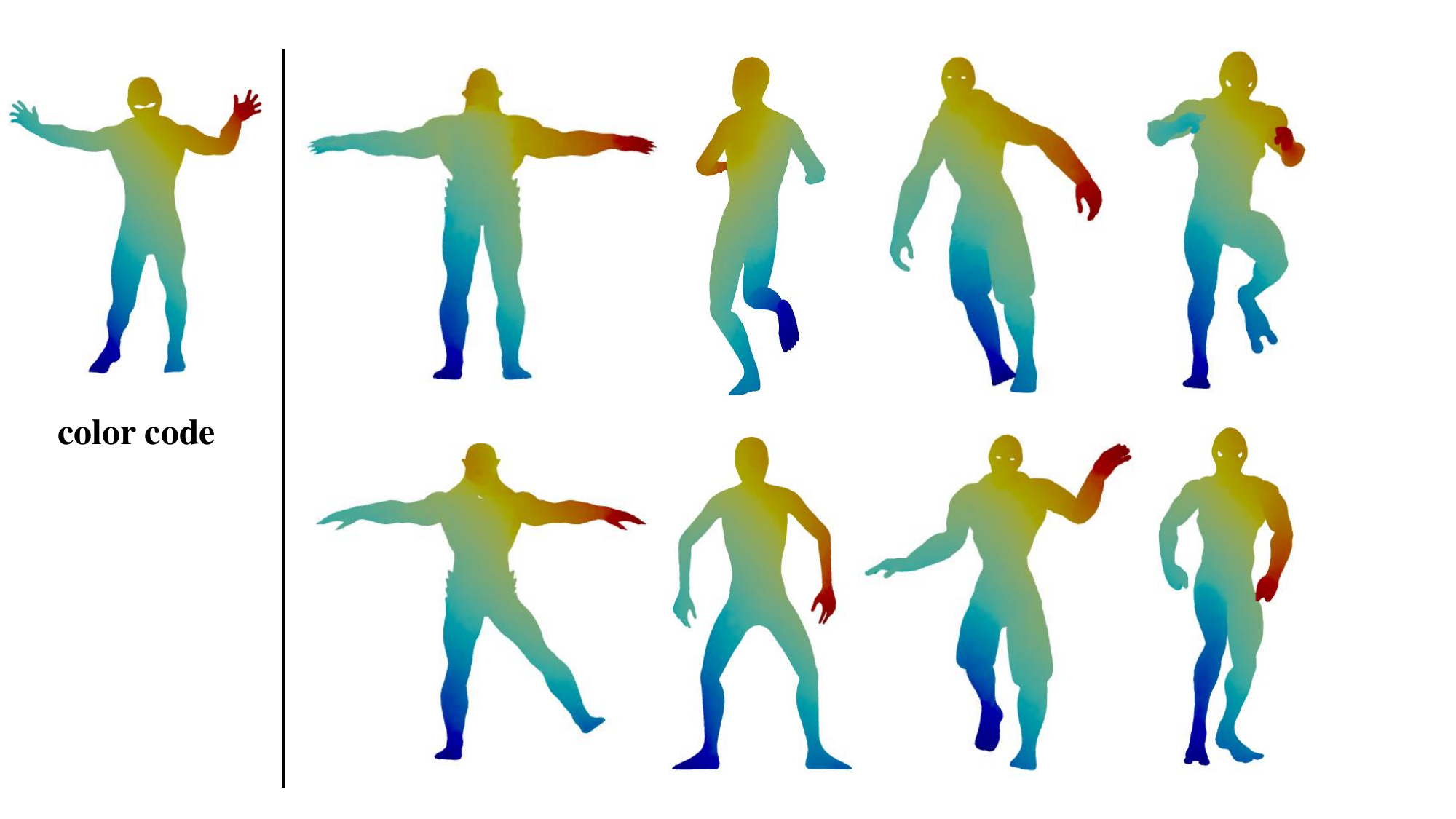}
\caption{\textbf{Qualitative examples on DT4D inter class dataset.}}
\label{fig:comparsion_DT4D}
\end{figure}

We provide qualitative examples of the correspondences produced by our method in Fig. \ref{fig:comparsion_FAUST}–\ref{fig:comparsion_DT4D}. As shown, our approach consistently generates smooth and semantically accurate correspondences across a wide range of shapes and poses, highlighting its robustness in handling complex non-rigid deformations.

\end{document}